\documentclass[letterpaper, 10pt, conference]{ieeeconf}      

\usepackage{times}
\usepackage{tabularx}
\usepackage{subcaption}
\usepackage{multirow,multicol, array}
\usepackage[font={small}]{caption}   
\usepackage{graphicx}
\usepackage{wrapfig}
\usepackage{siunitx}
\usepackage{soul}
\let\labelindent\relax
\usepackage{enumitem}

\usepackage{acronym}
\usepackage{balance}
\usepackage{mdwmath}
\usepackage{bm}
\usepackage{mdwtab}
\usepackage{eqparbox}
\usepackage{amssymb}
\usepackage{color}
\usepackage{xcolor}
\usepackage{psfrag}
\usepackage{epsfig}
\usepackage{epstopdf}
\usepackage{booktabs}
\usepackage{blindtext}
\usepackage{physics}
\usepackage{amsmath,amsfonts}
\usepackage{algorithmic}
\usepackage{algorithm}
\usepackage{array}
\usepackage[caption=false,font=normalsize,labelfont=sf,textfont=sf]{subfig}
\usepackage{textcomp}
\usepackage{stfloats}
\usepackage{url}
\usepackage{verbatim}
\usepackage{cite}
\usepackage{mathtools}
\newcommand{\defeq}{ \stackrel{\triangle}{=} } 
\newcommand{\tlcorner}[2]{\prescript{#1}{}{#2}} 
\newcommand{\trcorner}[2]{{#2}^{#1}} 
\newcommand{\blcorner}[2]{\prescript{}{#1}{#2}} 
\newcommand{\tblcorner}[3]{\prescript{#1}{#2}{#3}} 

\newcommand{\MassBLR}[2]{\blcorner{#1}{m}_{#2} } 

\newcommand{\ObjBLR}[3]{\blcorner{#2}{\mathbf{#1}}_{#3} } 
\newcommand{\ObjPlainBLR}[3]{\blcorner{#2}{#1}_{#3} } 
\newcommand{\ObjTildeBLR}[3]{\blcorner{#2}{\Tilde{\mathbf{#1}}}_{#3} } 
\newcommand{\ObjPrimeBLR}[3]{\blcorner{#2}{\mathbf{#1}}^{\prime}_{#3} } 

\newcommand{\CoMInertiaBLR}[2]{\blcorner{#1}{\mathbf{I}}^{cm}_{#2} } 

\newcommand{\Identity}[1]{\mathbf{1}_{#1 \times #1} } 

\newcommand{\BjMassBi}{\MassBLR{j}{i} }

\newcommand{\BjCoMInertiaBi}{\CoMInertiaBLR{j}{i} }

\newcommand{\BjObjBi}[1]{\ObjBLR{#1}{j}{i} }
\newcommand{\BjObjPlainBi}[1]{\ObjPlainBLR{#1}{j}{i} }
\newcommand{\BjObjTildeBi}[1]{\ObjTildeBLR{#1}{j}{i} }
\newcommand{\BjObjPrimeBi}[1]{\ObjPrimeBLR{#1}{j}{i} }

\makeatletter
\newcommand\fs@spaceruled{\def\@fs@cfont{\bfseries}\let\@fs@capt\floatc@ruled
  \def\@fs@pre{\vspace{0.5\baselineskip}\hrule height.8pt depth0pt \kern2pt}%
  \def\@fs@post{\kern2pt\hrule\vspace{-0.5\baselineskip}}
  \def\@fs@mid{\kern2pt\hrule\kern2pt}%
  \let\@fs@iftopcapt\iftrue}
\makeatother

\IEEEoverridecommandlockouts
\graphicspath{{images/}}

\title{
Deformable Multibody Modeling for Model Predictive Control in \\ Legged Locomotion with Embodied Compliance 
}
	
\author{Keran Ye and Konstantinos Karydis
	\thanks{The authors are with the Dept. of Electrical and Computer Engineering, University of California, Riverside. 
	Email: {\{kye007, karydis\}@ucr.edu}.
	}
	\thanks{
	We gratefully acknowledge the support of NSF under grant \#CMMI-2046270. Any opinions, findings, and conclusions or recommendations expressed in this material are those of the authors and do not necessarily reflect the views of the National Science Foundation.}
}
	
\begin{document}

\maketitle
\thispagestyle{empty}
\pagestyle{empty}


\begin{abstract}
The paper presents a method to stabilize dynamic gait for a legged robot with embodied compliance. Our approach introduces a unified description for rigid and compliant bodies to approximate their deformation and a formulation for deformable multibody systems. We develop the centroidal composite predictive deformed inertia (CCPDI) tensor of a deformable multibody system and show how to integrate it with the standard-of-practice model predictive controller (MPC). Simulation shows that the resultant control framework can stabilize trot stepping on a quadrupedal robot with both rigid and compliant spines under the same MPC configurations. Compared to standard MPC, the developed CCPDI-enabled MPC distributes the ground reactive forces closer to the heuristics for body balance, and it is thus more likely to stabilize the gaits of the compliant robot. A parametric study shows that our method preserves some level of robustness within a suitable envelope of key parameter values. 
\end{abstract}

\section{Introduction}
Benefiting from developments in actuation, sensory, and embedded computation, dynamic quadrupedal robot locomotion has made strides in terms of  agility~\cite{ijspeert2014biorobotics,hwangbo2019learning,park2017high,rudin2021cat,nguyen2019optimized,bledt2020extracting,wang2023unified} and robustness~\cite{katz2019mini,bledt2018cheetah,pandala2022robust,lee2019robust,lee2020learning,bellegarda2022robust} in mainstream designs~\cite{bouman2020autonomous,smith2022legged,katz2019mini,bledt2018cheetah,hutter2017anymal,ma2020bipedal,blackman2016gait,semini2015design,zucker2011optimization} where the main body is a single rigid body, and the overall morphological degree-of-freedoms (DoFs) are extended by legs. 
The main body's rigidly fixed morphology affords physically-induced simplifications that enable online model-based control strategies~\cite{di2018dynamic,bledt2019implementing,hutter2016anymal} to stabilize dynamic gaits. 
Nevertheless, further increasing morphological DoFs may be the key to extending maneuverability~\cite{koutsoukis2016passive,li2020trotting,fisher2017effect,zhang2016effects,liu2022systematic,ye2023evaluation,casarez2018steering}, with focus either on legs or the main body. 

Legs typically contain more DoFs, yet adding more actuation may be expensive~\cite{caporale2023twisting} and undermine the high-frequency response required for leg regulation. 
On the other hand, the main body holds the potential for more flexibility drawing from biological inspiration~\cite{zhang2014effect,schilling2010function,alexander1985elastic,schilling2011evolution}. 
A main body with increased morphological DoFs can be abstracted into spinal models. 
Vertebrate studies have justified the spinal functions such as twisting and bending~\cite{bennett2001twisting} in biological counterparts and identified the power source from a mixture of muscles (as actuators), and ligaments and tendons (as compliance)~\cite{schilling2010function,alexander1985elastic,schilling2011evolution}. 
Such studies motivate foundational robotics research to develop more sophisticated spine designs and associated control methods. 

\begin{figure}[!t]
	\vspace{6pt}
	\centering
\includegraphics[trim={0cm 0cm 0cm 0cm},clip,width=0.90\linewidth]{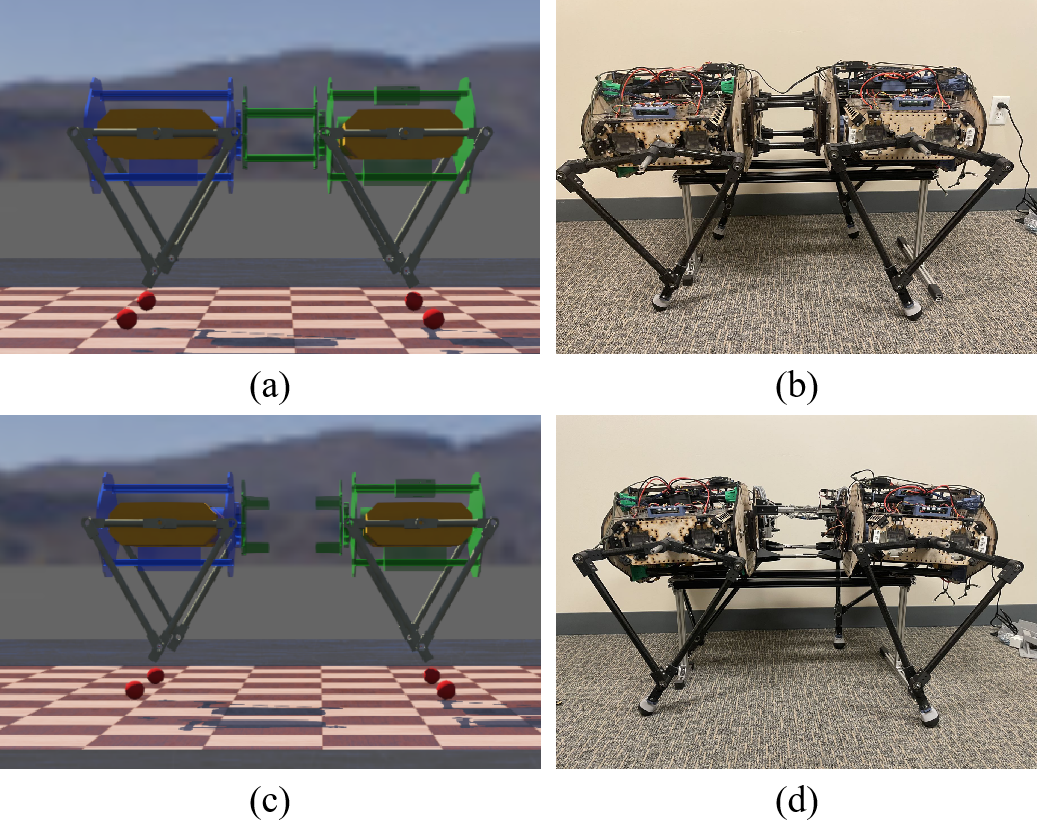}
	\vspace{-5pt}
	\caption{\textbf{Simulation Models and Robot Prototypes.} 
 (a) The simulation model with a rigid spine. 
 (b) The robot prototype with a rigid spine. 
(c) The simulation model with a compliant prismatic spine. 
 (d) The robot prototype with a compliant prismatic spine.  }\label{fig:rigid_and_compliant_models_prototypes}
	\vspace{-18pt}
\end{figure}

Increasing the actuated morphological DoFs in spines may enhance the overall controllability of a legged system. 
Indeed, approaches that have focused on integrating spines in legged robots have reported improvements in terms of gait stabilization~\cite{zhao2012embodiment,hustig2016morphological,takuma2010facilitating,eckert2015comparing,eckert2018towards,li2020trotting}, motion acceleration~\cite{folkertsma2012parallel,eckert2018towards,pusey2013free,bhattacharya2019learning}, and efficiency~\cite{eckert2018towards,haomachai2021lateral,bhattacharya2019learning}. 
However, the additional actuation comes with a major increase in weight which may challenge the overall payload that is mainly compensated by legs. 
In contrast to additional actuation, integrating compliance into the spine may introduce similar improvements while not increasing the weight by that much. 
Research efforts along this direction have identified benefits in improving energy efficiency~\cite{khoramshahi2013piecewise}, regulating gaits~\cite{koutsoukis2016passive,kani2011effect}, and enhancing agility~\cite{ye2021modeling}. 
However, integration of spinal compliance renders modeling and whole-body control more complex~\cite{ye2021modeling}.  

In the context of model-based control, the actuated morphological DoFs in the spine extend the main body to a rigid multibody system whose centroidal dynamics can be analyzed through efficient algorithms~\cite{siciliano2016robotics} and induced into predictive~\cite{katayama2023model,tazaki2024trajectory} or reactive controllers~\cite{orin2013centroidal,he2024cdm}. 
On the other hand, the compliant morphological DoFs turn the main body into a deformable multibody system which is considerably more complex. 
The exact deformation characterization of a compliant body is directly related to its compliant mechanism and requires a specific strain descriptor depending on the materials. 
Therefore, a general description of a deformable multibody system is lacking and its corresponding centroidal dynamics analysis remains an open area. 

This work aims to extract a unified description to represent legged robots with both rigid and compliant spines~\cite{ye2023novel} (Fig.~\ref{fig:rigid_and_compliant_models_prototypes}) and enhance model-based control. 
The major technical result concerns an embodied-compliance-aware model predictive controller (MPC) for quadrupedal gait stabilization. 
We focus on algorithm development and verification in simulation using a digital twin of our physical robot developed herein. 
The detailed contribution of this work includes 
(1) algorithms for approximating and predicting the centroidal properties of a deformable body and a deformable multibody system within a prediction horizon, 
(2) integration of the derived predictive centroidal properties into an MPC controller, and  
(3) simulation tests on a quadruped with various compliant settings for verifying the feasibility of the proposed strategy.

\section{Robot Model: Deformable Multibody System}

A dynamic legged robot is generally considered an underactuated multibody system that requires its kinodynamics to be first established for state estimation and dynamic control under mainstream whole-body regulation frameworks~\cite{di2018dynamic,dai2014whole}. 
A general model description rising from rigid multibody system theory~\cite{orin2013centroidal} has helped derive more specific models tailored to actual robot prototypes, including
\begin{itemize}
    \item single-body model~\cite{di2018dynamic}: It is suitable for robots with a rigid trunk and lightweight legs such that they can be assumed massless. 
    \item full-body model~\cite{posa2014direct}: It is a more accurate but complex way to incorporate the kinodynamics of each body to exploit the full dynamic range of a legged system. 
    \item centroidal model~\cite{orin2013centroidal}: It is a compressed approach projecting all bodies' information onto the robot's centroid and handling the instantaneous centroidal dynamics as the major interest of regulation and prediction. 
\end{itemize}

Introducing embodied compliance rejects the potato model by definition to represent the whole-body kinodynamics. 
On the other hand, we propose to embed compliance into the multibody concept with a deformable approximation notion such that system and centroidal quantities are traceable following the routines from the rigid multibody context.

\subsection{Deformable Body Configuration}
\label{subsec:deformable_body}
Describing a body object in terms of kinematics and dynamics properties is essential and fundamental in a multibody system. 
We revisit the rigid body description and then describe our proposed deformable body notion. 

For the $i^{th}$ rigid body, $\mathcal{B}_i$, body frame $\blcorner{0}{f}_i$ is used to define the body's Center of Mass (CoM) and Moment of Inertia (MoI); CAD model properties are typically used to deduce their specific expression. 
CoM and MoI are time-invariant under frame $\blcorner{0}{f}_i$. 
The inertia tensor is defined as
\begin{align}
\tlcorner{0}{\mathbf{I}}_i  
=
\begin{bmatrix}
  \mathbf{I}^{cm}_i + m_i \mathbf{S}(\mathbf{r}_i) \mathbf{S}(\mathbf{r}_i)^\top   
  & 
   m_i \mathbf{S}(\mathbf{r}_i)
   \\
    m_i \mathbf{S}(\mathbf{r}_i)^{\top}
    &
     m_i \Identity{3}
\end{bmatrix}\;,
\end{align}
where $m_i$ is the mass, $\mathbf{r}_i$ the CoM position expressed in frame $\blcorner{0}{f}_i$, and the symmetric matrix $\mathbf{I}^{cm}_i$ is the rotational inertia around the CoM and its principle axes are parallel to those of frame $\blcorner{0}{f}_i$.  
The body frame $\blcorner{0}{f}_i$ is treated as the only parent frame for the body $\mathcal{B}_i$ and it is used in the connection to the predecessor (or parent body). On the other hand, more child frames as $\blcorner{j}{f}_i$ could be used to link to the successors. 

In the context of continuum mechanics, the morphology of a deformable body can be described with the deformation mapping and the further deduced deformation gradient and strain tensor. 
The deformation mapping specifies every particle within the same body from the current position to the next position. 
The collection of all particle positions is denoted as a configuration. 
In this way, the instantaneous CoM and MoI can be defined from the current configuration to the next configuration using the deformation mapping. 
Nevertheless, the deformation mapping may not be available in real-time for online computation. 

\begin{figure}[!t]
	\vspace{6pt}
\includegraphics[trim={0cm 0cm 0cm 0cm},clip,width=\linewidth]{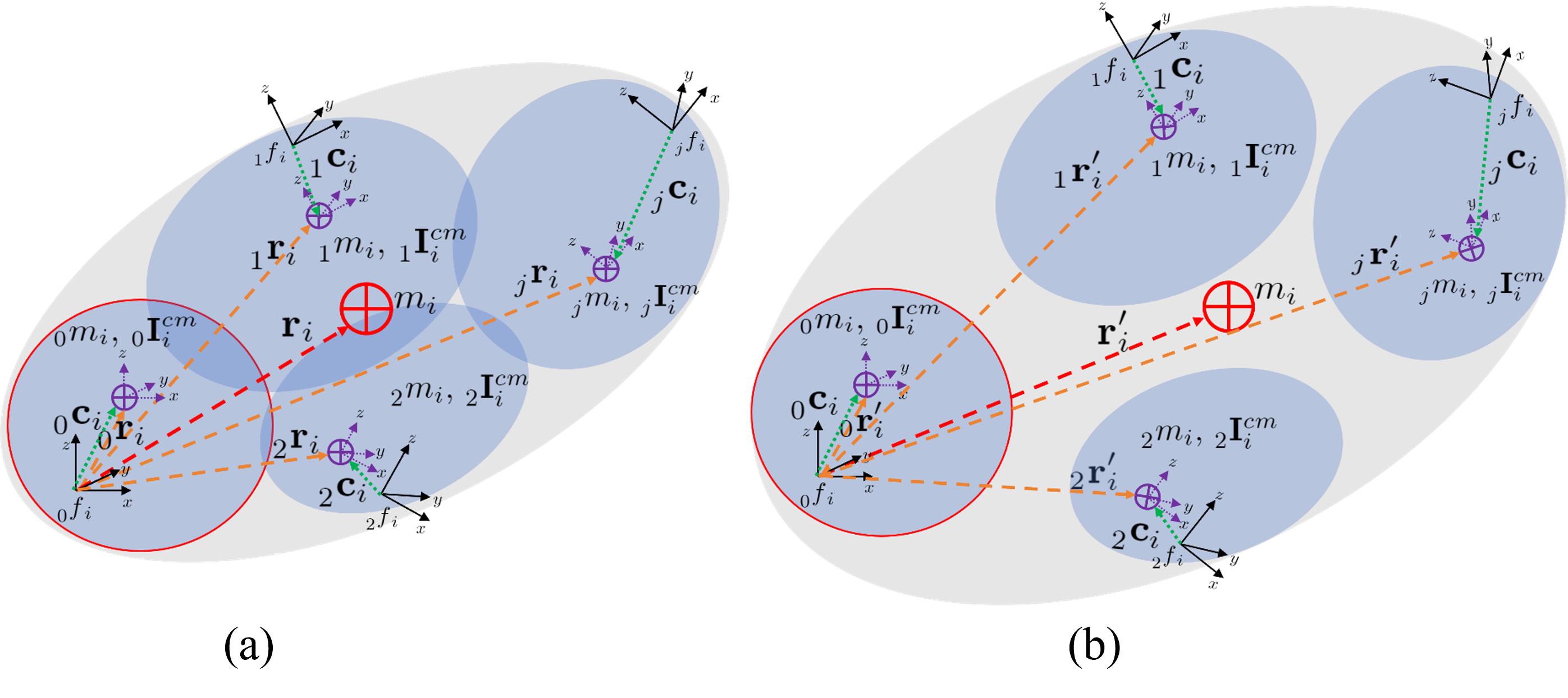}
	\vspace{-18pt}
	\caption{\textbf{Deformable Body Approximation.} 
 (a) The initial configuration of a body (grey): each sub-body $\mathcal{B}_i$ (blue) has a local frame $\blcorner{j}{f}_i$, a CoM (purple cross-circle) and a MoI (purple axes). The Sub-body CoM position vector is denoted as $\BjObjBi{c}$ (green dashed arrow) in the frame $\blcorner{j}{f}_i$ and $\BjObjBi{r}$ (orange dashed arrow) in the frame $\blcorner{0}{f}_i$. The body has an overall CoM (red cross-circle) and its position vector is denoted as $\mathbf{r}_i$ (red dashed arrow) in the frame $\blcorner{0}{f}_i$.
 (b) The deformed configuration of a body: sub-bodies move relative to each other with their CoM position vectors denoted as $\BjObjPrimeBi{r}$ (orange dashed arrow). The body CoM is denoted as $\mathbf{r}_i^{\prime}$ (red dashed arrow). 
 }\label{fig:model_deformable_body_summary}
	\vspace{-18pt}
\end{figure}

We develop an approximation of the deformation mapping using several body frames (Fig.~\ref{fig:model_deformable_body_summary}). 
We assume a deformable body  $\mathcal{B}_i$ can be decomposed into $N_i$ sub-bodies $\blcorner{j}{\mathcal{B}}_i$ with the left-subscript $\blcorner{j}{}$ denoting the sub-body index. 
The sub-body $\blcorner{j}{\mathcal{B}}_i$ is mostly a rigid cluster of particles and deformation is indicated through relative motion between sub-bodies.  Each sub-body  $\blcorner{j}{\mathcal{B}}_i$ possesses a body frame $\blcorner{j}{f}_i$ like that in a rigid body, such that the sub-body $\blcorner{0}{\mathcal{B}}_i$ is with the parent frame $\blcorner{0}{f}_i$ and the other sub-bodies with the child frames.

Each sub-body $\blcorner{j}{\mathcal{B}}_i$ has known mass, $\blcorner{j}{m}_i$, CoM position, $\blcorner{j}{\mathbf{c}}_i$, and rotational inertia, $\blcorner{j}{\mathbf{I}}^{cm}_i$, about the CoM expressed in frame $\blcorner{j}{f}_i$.   
Therefore, its instantaneous inertia tensor can be expressed in frame $\blcorner{j}{f}_i$ as
\begin{align}
\blcorner{j}{\mathbf{I}}_i 
= 
\begin{bmatrix}
\blcorner{j}{\mathbf{I}}^{cm}_i  
+ 
\blcorner{j}{m}_i \mathbf{S}(\blcorner{j}{\mathbf{c}}_i) \mathbf{S}(\blcorner{j}{\mathbf{c}}_i)^\top 
& 
\blcorner{j}{m}_i \mathbf{S}(\blcorner{j}{\mathbf{c}}_i) 
\\
\blcorner{j}{m}_i \mathbf{S}(\blcorner{j}{\mathbf{c}}_i)^\top 
& 
\blcorner{j}{m}_i \mathbf{1}_{3 \times 3}
\end{bmatrix}\;;
\label{equ:subbody_insta_inertia_subbody_frame_def}
\end{align}
$\mathbf{S}(\cdot)$ is the skew-symmetric operator and $\blcorner{j}{\mathbf{I}}_i$ is time-invariant. 

Between sub-bodies, we are interested in kinematics of the child frames $\blcorner{j}{f}_i$  with respect to the parent frame  $\blcorner{0}{f}_i$ such that the resultant properties are expressed in the frame $\blcorner{0}{f}_i$ and consistent with the rigid body representation. The sub-body kinematics relations are defined as
\begin{align}
\blcorner{j}{\mathbf{T}}_i
=
\begin{bmatrix}
\blcorner{j}{\mathbf{R}}_i & \blcorner{j}{\mathbf{p}}_i \\
\mathbf{0}_{1 \times 3} & 1
\end{bmatrix} 
,
\BjObjBi{X}
=
\textit{Ad}(\BjObjBi{T})
,
\blcorner{j}{\mathbf{V}}_i
=
\begin{bmatrix}
\blcorner{j}{\boldsymbol{\omega}}^\top_i, \blcorner{j}{\mathbf{v}}^\top_i
\end{bmatrix}^\top \hspace{-3pt}.
\label{equ:instantaneous_subbody_kinematics}
\end{align}
$\blcorner{j}{\mathbf{T}}_i$ and $\BjObjBi{X}$ are the instantaneous sub-body transformation and adjoint transformation of frame $\blcorner{j}{f}_i$ relative to frame $\blcorner{0}{f}_i$. 
$\blcorner{j}{\mathbf{V}}_i$ is the sub-body twist of frame $\blcorner{j}{f}_i$'s origin relative to frame $\blcorner{0}{f}_i$ expressed in the frame $\blcorner{j}{f}_i$. 
Rotation $\blcorner{j}{\mathbf{R}}_i$, translation $\blcorner{j}{\mathbf{p}}_i$, angular velocity $\blcorner{j}{\boldsymbol{\omega}}_i$, and linear velocity $\blcorner{j}{\mathbf{v}}_i$ are also defined for frame $\blcorner{j}{f}_i$ relative to frame $\blcorner{0}{f}_i$. 
The sub-body inertia tensor from~\eqref{equ:subbody_insta_inertia_subbody_frame_def} can be expressed in frame $\blcorner{0}{f}_i$ and composed to the instantaneous inertia tensor, $\tlcorner{0}{\mathbf{I}}_i$, as
\begin{align}
\tblcorner{0}{j}{\mathbf{I}}_i 
= 
\BjObjBi{X}^{-\top}
\blcorner{j}{\mathbf{I}}_i
\BjObjBi{X}^{-1}
, \;
\tlcorner{0}{\mathbf{I}}_i 
&= 
\sum_{j=0}^{N_i-1} \tblcorner{0}{j}{\mathbf{I}}_{i}\;,
\label{equ:body_insta_inertia_def}
\end{align}
where $N_i$ is the number of sub-bodies in the body $\mathcal{B}_i$. 

\textbf{Remark 1:} 
Tensor $\tblcorner{0}{j}{\mathbf{I}}_i$ only captures the dynamical properties at the moment without any implication on its change tendency in the future. 
Thus, we are interested in predicting the future CoM and inertia accounting for body deformation along the prediction horizon in the MPC control paradigm. 

We propose to approximate the body inertia tensor after deformation with the predictive deformed inertia (PDI) tensor $\tlcorner{0}{\mathbf{I}}^{(k)}_{i}$ based on step time $\Delta t$, step count $k$, and the current sub-body twist $\blcorner{j}{\mathbf{V}}_i$ from~\eqref{equ:instantaneous_subbody_kinematics}. 
The predicted properties of each sub-body $\blcorner{j}{\mathcal{B}}_i$ are first investigated, followed by the composition of the whole body. 

Frame $\blcorner{j}{f}_{i}^{(k)}$ refers to the predicted sub-body frame at the $k$ step; hence, $\blcorner{j}{f}_{i}^{(0)}=\blcorner{j}{f}_{i}$. 
Figure~\ref{fig:model_deformable_body_summary} demonstrates the trajectory of frame $\blcorner{j}{f}_{i}^{(k)}$ when the sub-body twist $\blcorner{j}{\mathbf{V}}_i$ is time-invariant along the whole prediction horizon. 
The kinematics of frame $\blcorner{j}{f}_{i}^{(k)}$ relative to $\blcorner{j}{f}_{i}^{(k-1)}$ can be characterized with the incremental sub-body transformation $\blcorner{j}{\Tilde{\mathbf{T}}}_i$ and its adjoint transformation $\BjObjTildeBi{X}$ in the exponential coordinates~\cite{lynch2017modern} as
\begin{align}
\blcorner{j}{\Tilde{\mathbf{T}}}_i 
\defeq 
\mathbf{e}^{ [ \BjObjBi{S} ] \BjObjPlainBi{\theta} } 
, 
\BjObjTildeBi{X}
=
\textit{Ad}(\BjObjTildeBi{T})\;,
\label{equ:incremental_subbody_adjoint_transformation}
\end{align}
where $ \BjObjBi{S}$ is the screw axis and $\blcorner{j}{\theta}_{i}$ the incremental travel distance that can be computed based on the sub-body twist $\blcorner{j}{\mathbf{V}}_i$ in a closed-formed expression~\cite{lynch2017modern}. 

The time invariance of $\BjObjBi{I}$ suggests that it stays the same after the sub-body frame $\blcorner{j}{f}_{i}$ moves in $k$ steps; in other words, $\BjObjBi{I}^{(k)}=\BjObjBi{I}$ expressed in frame $\blcorner{j}{f}_{i}^{(k)}$. 
It can be transformed backward into frame $\blcorner{0}{f}_{i}$ via
\begin{align}
\tblcorner{0}{j}{\mathbf{I}}_{i}^{(k)}
=
\BjObjBi{Y}^{(k)\top} \BjObjBi{I}^{(k)} \BjObjBi{Y}^{(k)}
=
\BjObjBi{Y}^{(k)\top} \BjObjBi{I} \BjObjBi{Y}^{(k)}\;,
\label{equ:onestep_predictive_subbody_inertia_def}
\end{align}
where $\blcorner{j}{\mathbf{Y}}_{i}^{(k)}$ is the predictive adjoint transformation (PAT) in $k$ steps and can be calculated iteratively via
\begin{equation}
\hspace{-3pt}\BjObjBi{Y}^{(k)} = 
\begin{cases}
&\BjObjTildeBi{Y} \BjObjBi{Y}^{(k-1)}
, 
\BjObjTildeBi{Y} 
=
\BjObjTildeBi{X}^{-1}\;,\textit{ for } k>0
\\
&\BjObjBi{X}^{-1}\;,\textit{ for } k=0
\label{equ:predictive_subbody_adjoint_transformation_iterative}
\end{cases},
\end{equation}
where $\BjObjTildeBi{Y}$ is the predictive incremental adjoint transformation (PIAT) in one step. 
The PDI tensor in $k$ steps can be derived through summation
\begin{align}
\tlcorner{0}{\mathbf{I}}_{i}^{(k)} 
= 
\sum_{j=0}^{N_i-1} \tblcorner{0}{j}{\mathbf{I}}_{i}^{(k)}
=
\begin{bmatrix}
\bar{\mathbf{I}}_i^{(k)} 
& 
m_i \mathbf{S}(\mathbf{r}_i^{(k)}  ) 
\\
m_i \mathbf{S}(\mathbf{r}_i^{(k)}  )^\top 
& 
m_i \mathbf{1}_{3 \times 3}
\end{bmatrix}\; .
\label{equ:body_predictive_deformed_inertia_def}
\end{align}

\textbf{Remark 2:} 
From~\eqref{equ:predictive_subbody_adjoint_transformation_iterative} we see that $\BjObjTildeBi{Y}$ is the adjoint transformation of frame $\blcorner{j}{f}_{i}^{(k-1)}$ relative to $\blcorner{j}{f}_{i}^{(k)}$ , and  $\BjObjBi{Y}^{(k)}$ is the adjoint transformation of frame $\blcorner{j}{f}_{i}^{(0)}$ relative to $\blcorner{j}{f}_{i}^{(k)}$. 

The predictive adjoint transformation $\BjObjBi{Y}^{(k)}$ characterizes the partial deformation from sub-body $\blcorner{j}{\mathcal{B}}_i$; how to be exploited in centroidal composite predictive deformed inertia is discussed in Section~\ref{subsec:centroidal_inertia}. 
When $k=0$, the PDI tensor $\tlcorner{0}{\mathbf{I}}_{i}^{(k)}$ degrades to the instantaneous body inertia as $\tlcorner{0}{\mathbf{I}}_{i} = \tlcorner{0}{\mathbf{I}}_{i}^{(0)}$. 
The pseudo-coded algorithm of the PDI tensors up to the horizon count $N_p - 1$ is summarized in Alg.~\ref{alg:deformable_body_inertia}. 
\vspace{-12pt}

\floatstyle{spaceruled}
\restylefloat{algorithm}
\begin{algorithm}[H]
\caption{Computing Predictive Deformed Inertia Tensors and Predictive Adjoint Transformation of Single Deformable Body $\mathcal{B}_i$.}\label{alg:deformable_body_inertia}
\begin{algorithmic}[1]
\STATE {\textbf{Inputs}: }  $\blcorner{j}{\mathbf{T}}_i$,  $\blcorner{j}{\mathbf{V}}_i$, $\BjObjBi{X}$
\STATE {\textbf{Outputs}: }  $\tlcorner{0}{\mathbf{I}}_{i}^{(k)}$,  $\BjObjBi{Y}^{(k)}$  
\STATE {\textbf{Model data}: } $N_i$, $\BjMassBi$, $\BjCoMInertiaBi$,  $\blcorner{j}{\mathbf{c}}_i$, $\Delta t$, $N_p$  
\STATE \hspace{0.0cm} \textbf{for} $j=0$ to $N_i - 1$ \textbf{do}
\STATE \hspace{0.3cm} compute $\blcorner{j}{\mathbf{I}}_i$ and $\BjObjTildeBi{X}$ via~\eqref{equ:subbody_insta_inertia_subbody_frame_def} and~\eqref{equ:incremental_subbody_adjoint_transformation}
\STATE \hspace{0.3cm} compute $\BjObjTildeBi{Y} = \BjObjTildeBi{X}^{-1}$
\STATE \hspace{0.3cm} initialize $\BjObjBi{Y}^{(0)}=\BjObjBi{X}^{-1}$
\STATE \hspace{0.3cm} \textbf{for} $k=0$ to $N_p - 1$ \textbf{do}
\STATE \hspace{0.6cm} $\BjObjBi{Y}^{(k+1)}= \BjObjTildeBi{Y} \BjObjBi{Y}^{(k)}$ 
\STATE \hspace{0.6cm} $\tblcorner{0}{j}{\mathbf{I}}_{i}^{(k)} = \BjObjBi{Y}^{(k)\top} \BjObjBi{I} \BjObjBi{Y}^{(k)}$
\STATE \hspace{0.3cm} \textbf{end}
\STATE \hspace{0.0cm} \textbf{end}
\STATE \hspace{0.0cm} \textbf{for} $k=0$ to $N_p - 1$ \textbf{do}
\STATE \hspace{0.3cm} $\tlcorner{0}{\mathbf{I}}_{i}^{(k)} = \sum_{j=0}^{N_i-1} \tblcorner{0}{j}{\mathbf{I}}_{i}^{(k)}$
\STATE \hspace{0.0cm} \textbf{end}
\end{algorithmic}
\label{alg1}
\end{algorithm}

\textbf{Remark 3:} 
The deformation of body $\mathcal{B}_i$ is indicated by the sub-body twist $\BjObjBi{V}$. 
Therefore, rigidness implies  $\BjObjBi{V} = \mathbf{0}_{6 \times 1}$ for any $j$. 
This suggests that the proposed deformable body notion is a unified description that can model a rigid body by setting all sub-body twist quantities to zero.

\subsection{Kinematics Propagation}

\begin{figure}[!t]
	\vspace{6pt}
\includegraphics[trim={0cm 0cm 0cm 0cm},clip,width=\linewidth]{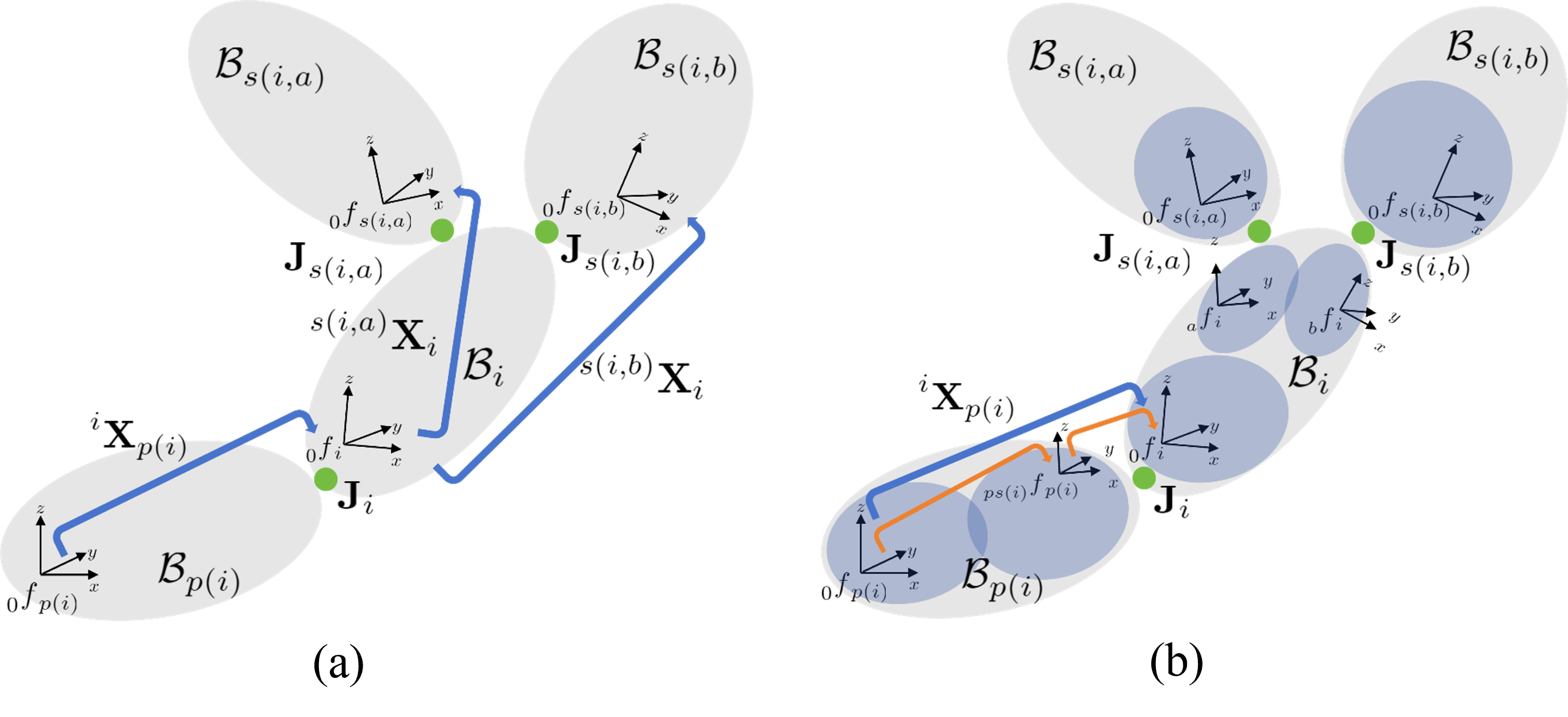}
	\vspace{-18pt}
	\caption{\textbf{Multibody Representation.} 
 (a) The rigid multibody tree topology: each body (grey) is connected to its preceding body through a joint (green). The adjoint transform $\tlcorner{i}{\mathbf{X}}_{p(i)}$ (blue arrow) is defined by the joint $\mathbf{J}_i$.  
 (b) The deformable multibody tree topology: each body is linked to its preceding body through a joint and a corresponding sub-body (blue). The adjoint transform $\tlcorner{i}{\mathbf{X}}_{p(i)}$ (blue arrow) is defined by the joint $\mathbf{J}_i$ and deformation (orange arrow). 
 }\label{fig:model_multibody_summary}
	\vspace{-15pt}
\end{figure}

The tree structure is one of the networks to represent the rigid multibody system~\cite{siciliano2016robotics}. 
Figure~\ref{fig:model_multibody_summary}(a) illustrates such linking relations between adjacent bodies that each body is connected to its predecessor with a joint. 
One of the bodies is treated as the root. The root body is also a floating base and possesses a fictitious joint to the world frame~\cite{siciliano2016robotics}. 

We extend this topology with the proposed deformable body representation, with the core idea being that each body is connected to its predecessor with a joint and a corresponding sub-body from the predecessor, as shown in Fig.~\ref{fig:model_multibody_summary}(b). 
Let body $\mathcal{B}_i$ be the current body. 
Then, the predecessor is $\mathcal{B}_{p(i)}$~\cite{orin2013centroidal}. 
When the predecessor $\mathcal{B}_{p(i)}$ is deformable, its sub-body $\mathcal{B}_{p(i),ps(i)}$ connects to $\mathcal{B}_i$. 
When $\mathcal{B}_i$ is deformable, its $j^{th}$ sub-body links to the successor $\mathcal{B}_{s(i,j)}$. 
Let $\mathcal{B}_0$ be the root body with major frame $\blcorner{0}{f}_0$.



The adjoint transformation between bodies $\mathcal{B}_i$ and $\mathcal{B}_{p(i)}$ is essentially between the major frames $\blcorner{0}{f}_{p(i)}$ and $\blcorner{0}{f}_{i}$. 
Their twist propagation is expressed as
\begin{align}
\tlcorner{i}{\mathbf{V}}_{i}
=
\tlcorner{i}{\mathbf{X}}_{p(i)} \tlcorner{p(i)}{\mathbf{V}}_{p(i)}
+
\tlcorner{i}{\mathbf{V}}_{i,p(i)}\;,
\label{equ:general_twist_propagation}
\end{align}
where $\tlcorner{i}{\mathbf{X}}_{p(i)}$ is the adjoint transformation of $\blcorner{0}{f}_{p(i)}$ relative to $\blcorner{0}{f}_{i}$ and $\tlcorner{i}{\mathbf{V}}_{i,p(i)}$ is their relative twist expressed in $\blcorner{0}{f}_{i}$.  

We first revisit the rigid multibody case~\cite{orin2013centroidal}, as shown in Fig.~\ref{fig:model_multibody_summary}(a). 
The adjoint transformation and the relative twist in~\eqref{equ:general_twist_propagation} are solely defined through joint $\mathbf{J}_i$ described by its pose and velocity variables $\mathbf{q}_i$ and $\dot{\mathbf{q}}_i$~\cite{siciliano2016robotics,orin2013centroidal} as 
\begin{align}
\tlcorner{i}{\mathbf{X}}_{p(i)}
=
\tlcorner{i}{\mathbf{X}}_{p(i)} (\mathbf{q}_i)
, 
\tlcorner{i}{\mathbf{V}}_{i,p(i)}
=
\mathbf{\Phi}_i \dot{\mathbf{q}}_i\;,
\label{equ:rigid_twist_propagation_relative_twist_def}
\end{align}
where $\mathbf{\Phi}_i$ denotes the free modes of joint $\mathbf{J}_i$~\cite{siciliano2016robotics}. 

On the other hand, when body $\mathcal{B}_{p(i)}$ is deformable, the above twist propagation needs to be described in $k$ steps and can be decomposed into two phases. 
The first phase is from frame $\blcorner{0}{f}_{p(i)}$ to $\blcorner{ps(i)}{f}_{p(i)}^{(k)}$ via
\begin{align}
\tlcorner{ps(i)}{\mathbf{V}}_{ps(i)}^{(k)}
&=
\tlcorner{ps(i)}{\mathbf{X}}_{p(i)}^{(k)} \tlcorner{p(i)}{\mathbf{V}}_{p(i)}
+
\tlcorner{ps(i)}{\mathbf{V}}_{ps(i),p(i)}^{(k)} 
\nonumber
\\
&=
\blcorner{ps(i)}{\mathbf{Y}}_{p(i)}^{(k)}  \tlcorner{p(i)}{\mathbf{V}}_{p(i)} + \blcorner{ps(i)}{\mathbf{V}}_{p(i)}\;,
\label{equ:twist_propagation_deformable_first}
\end{align}
and the second phase is from frame $\blcorner{ps(i)}{f}_{p(i)}^{(k)}$ to $\blcorner{0}{f}_{i}$ via
\begin{align}
\tlcorner{i}{\mathbf{V}}_{i}
&=
\tlcorner{i}{\mathbf{X}}_{ps(i)}^{(k)} \tlcorner{ps(i)}{\mathbf{V}}_{ps(i)}^{(k)}
+
\tlcorner{i}{\mathbf{V}}_{i,ps(i)}^{(k)} 
\nonumber
\\
&=
\tlcorner{i}{\mathbf{X}}_{ps(i)}(\mathbf{q}_i) \tlcorner{ps(i)}{\mathbf{V}}_{ps(i)}^{(k)} 
+ 
\mathbf{\Phi}_i \dot{\mathbf{q}}_i\;.
\label{equ:twist_propagation_deformable_second}
\end{align}
Combining~\eqref{equ:twist_propagation_deformable_first} and~\eqref{equ:twist_propagation_deformable_second} yields
 \begin{align}
\tlcorner{i}{\mathbf{V}}_{i}
=&
\tlcorner{i}{\mathbf{X}}_{ps(i)}(\mathbf{q}_i) \, \blcorner{ps(i)}{\mathbf{Y}}_{p(i)}^{(k)}  \tlcorner{p(i)}{\mathbf{V}}_{p(i)} 
\nonumber
\\
&+
\tlcorner{i}{\mathbf{X}}_{ps(i)}(\mathbf{q}_i) \,\blcorner{ps(i)}{\mathbf{V}}_{p(i)} 
+
\mathbf{\Phi}_i \dot{\mathbf{q}}_i\;,
 \end{align}
which indicates that the adjoint transformation and the relative twist in~\eqref{equ:general_twist_propagation} are defined through the states of joint $\mathbf{J}_i$ and the predictive sub-body kinematics in $k$ steps as
\begin{align}
\tlcorner{i}{\mathbf{X}}_{p(i)}^{(k)}
&=
\tlcorner{i}{\mathbf{X}}_{ps(i)}(\mathbf{q}_i) \, \blcorner{ps(i)}{\mathbf{Y}}_{p(i)}^{(k)}\;,
\label{equ:deformable_twist_propagation_adjoint_transformation_def}
\\
\tlcorner{i}{\mathbf{V}}_{i,p(i)}
&=
\tlcorner{i}{\mathbf{X}}_{ps(i)}(\mathbf{q}_i) \,\blcorner{ps(i)}{\mathbf{V}}_{p(i)} 
+
\mathbf{\Phi}_i \dot{\mathbf{q}}_i\;,
\label{equ:deformable_twist_propagation_relative_twist_def}
\end{align}
where $\tlcorner{i}{\mathbf{X}}_{ps(i)}(\mathbf{q}_i)$, $\blcorner{ps(i)}{\mathbf{V}}_{p(i)}$ and $\mathbf{\Phi}_i \dot{\mathbf{q}}_i$ are deduced from joint and deformation measurements and $\blcorner{ps(i)}{\mathbf{Y}}_{p(i)}^{(k)}$ is computed in Alg.~\ref{alg:deformable_body_inertia}. 



\subsection{Centroidal Composite Predictive Deformed Inertia}
\label{subsec:centroidal_inertia}
The centroidal dynamics of a complex multibody system can be used for regulation when the overall motion is prioritized over a subordinate body motion~\cite{siciliano2016robotics}.
In the context of legged locomotion, the centroidal momentum can be exploited to design feedback control laws for reactive posture regulation~\cite{orin2013centroidal}. 
However, computation of the centroidal momentum usually requires to identify the centroidal inertia tensor first. 
The successful practice of model predictive control on legged platforms~\cite{bledt2018cheetah,hutter2017anymal,chignoli2021humanoid} also justified the explicit use of the centroidal inertia tensor to construct dynamical constraints. 
Therefore, we focus on the centroidal inertia tensor for a deformable multibody system. 

The literature regarding rigid multibody systems explores the centroidal composite rigid body inertia (CCRBI) as the centroidal inertia tensor of interest~\cite{orin2013centroidal,siciliano2016robotics}. 
An efficient algorithm~\cite{orin2013centroidal} exists to recursively compose the inertia tensor from the most remote body to the root body via
\begin{align}
\trcorner{C}{\mathbf{I}}_{p(i)} 
= 
\trcorner{C}{\mathbf{I}}_{p(i)}
+
\tlcorner{i}{\mathbf{X}}_{p(i)}^\top \trcorner{C}{\mathbf{I}}_{i} \tlcorner{i}{\mathbf{X}}_{p(i)}\;,
\end{align}
where $\trcorner{C}{\mathbf{I}}_{i}$ is the composite inertia tensor for the body $\mathcal{B}_i$ and its successors expressed in its major frame $\blcorner{0}{f}_{i}$. 
This indicates that the current body and its successors move as one to emphasize the overall motion. 
This recursive composition yields $\trcorner{C}{\mathbf{I}}_{0}$, the composite inertia tensor of the overall multibody system expressed in root-body frame $\blcorner{0}{f}_{0}$. 

The composition of rigid bodies can be extended to our proposed deformable bodies by considering the predictive deformed inertia tensor and predictive sub-body adjoint transformation. 
 The recursive composition at the prediction instant $k \Delta t$ is analogous to the rigid case and results in
 \begin{align}
&\trcorner{C,(k)}{\mathbf{I}}_{p(i)}
=
\trcorner{C,(k)}{\mathbf{I}}_{p(i)}
+
 (\tlcorner{i}{\mathbf{X}}_{p(i)}^{(k)})^{\top} 
 \; \trcorner{C,(k)}{\mathbf{I}}_{i} 
 \; \tlcorner{i}{\mathbf{X}}_{p(i)}^{(k)}\;,
\label{equ:recursive_deformable_composite_inertia_tensor}
\\
&\tlcorner{0}{\mathbf{I}}_{\mathcal{G}}^{(k)}
\defeq \
\trcorner{C,(k)}{\mathbf{I}}_{0}\;,
\label{equ:centroidal_composite_predictive_deformed_inertia_def}
\end{align}
where $\tlcorner{i}{\mathbf{X}}_{p(i)}^{(k)}$ is defined in~\eqref{equ:deformable_twist_propagation_adjoint_transformation_def} and $\trcorner{C,(k)}{\mathbf{I}}_{i} (k \Delta t)$ is the predictive deformed composite inertia (PDCI) tensor in $k$ steps and is initialized with the PDI tensor $\tlcorner{0}{\mathbf{I}}^{(k)}_i$ from Alg.~\ref{alg:deformable_body_inertia}. 
The centroidal composite predictive deformed inertia (CCPDI) tensor $\tlcorner{0}{\mathbf{I}}_{\mathcal{G}}^{(k)}$ is  expressed in $\blcorner{0}{f}_{0}$ and contains the predictive composite centroid vector $\tlcorner{0}{\mathbf{r}}_{\mathcal{G}}^{(k)}$. 

The center-body frame $f_{\mathcal{C}}$ has its origin anchored at the composite centroid and is a pure translation $\tlcorner{0}{\mathbf{p}}_{\mathcal{C}} = \tlcorner{0}{\mathbf{r}}_{\mathcal{G}}^{(k)}$ from the root-body major frame $\blcorner{0}{f}_0$ with the corresponding predicted adjoint transformation and CCPDI in $k$ steps as
\begin{align}
\tlcorner{0}{\mathbf{X}}_{\mathcal{C}}^{(k)}
&=
\begin{bmatrix}
\Identity{3} 
& 
\mathbf{0}_{3 \times 3} 
\\
\mathbf{S}( \tlcorner{0}{\mathbf{r}}_{\mathcal{G}}^{(k)} )
& 
\Identity{3} 
\end{bmatrix}\;,
\label{equ:center_root_adjoint_transformation}
\\
\tlcorner{\mathcal{C}}{\mathbf{I}}_{\mathcal{G}}^{(k)}
&= 
(\tlcorner{0}{\mathbf{X}}_{\mathcal{C}}^{(k)} )^{\top}
\tlcorner{0}{\mathbf{I}}_{\mathcal{G}}^{(k)} 
\tlcorner{0}{\mathbf{X}}_{\mathcal{C}}^{(k)}\;,
\label{equ:centroidal_composite_predictive_deformed_inertia_in_centerbody_frame}
\end{align}
and the instantaneous center-body twist is propagated via
\begin{align}
\tlcorner{\mathcal{C}}{\mathbf{V}}_{\mathcal{C}}
=
(\tlcorner{0}{\mathbf{X}}_{\mathcal{C}}^{(0)})^{-1} \, \tlcorner{0}{\mathbf{V}}_{0}\;,
\label{equ:center_body_twist_propagation}
\end{align}
where $\tlcorner{0}{\mathbf{V}}_{0}$ is the root-body twist expressed in the frame $\ObjPlainBLR{f}{0}{0}$ and can be estimated in the whole-body estimator. 
Algorithm~\ref{alg:deformable_multibody_centroidal_composite_inertia} summarizes the computation of the CCPDI tensor.

\floatstyle{spaceruled}
\restylefloat{algorithm}
\begin{algorithm}[!t]
\caption{Centroidal Composite Predictive Deformed Inertia Tensors of A Deformable Multibody System.}\label{alg:deformable_multibody_centroidal_composite_inertia}
\begin{algorithmic}
\STATE {\textbf{Inputs}: } $\tlcorner{0}{\mathbf{I}}_{i}^{(k)}$, $\BjObjBi{Y}^{(k)}$, $\tlcorner{i}{\mathbf{X}}_{ps(i)}(\mathbf{q}_i)$
\STATE {\textbf{Outputs}: } $\tlcorner{0}{\mathbf{I}}_{\mathcal{G}}^{(k)}$ , $\tlcorner{\mathcal{C}}{\mathbf{I}}_{\mathcal{G}}^{(k)}$
\STATE {\textbf{Model data}: } $N$, $N_p$
\STATE {\textbf{Initialization}: } 
\STATE \hspace{0.0cm} \textbf{for} $i=0$ to $N - 1$ \textbf{do}
\STATE \hspace{0.3cm} $\trcorner{C,(k)}{\mathbf{I}}_{i} = \tlcorner{0}{\mathbf{I}}^{(k)}_i$
\STATE \hspace{0.0cm} \textbf{end}
\STATE {\textbf{CCPDI Tensors in Root and Center-Body Frames}: }
\STATE \hspace{0.0cm} \textbf{for} $k=0$ to $N_p-1$ \textbf{do}
\STATE \hspace{0.3cm} \textbf{for} $i=N-1$ to $1$ \textbf{do}
\STATE \hspace{0.6cm} retrieve $\blcorner{ps(i)}{\mathbf{Y}}_{p(i)}^{(k)}$ from all  $\BjObjBi{Y}^{(k)}$
\STATE \hspace{0.6cm} $\tlcorner{i}{\mathbf{X}}_{p(i)}^{(k)}=\tlcorner{i}{\mathbf{X}}_{ps(i)}(\mathbf{q}_i) \, \blcorner{ps(i)}{\mathbf{Y}}_{p(i)}^{(k)}$
\STATE \hspace{0.6cm} $\trcorner{C,(k)}{\mathbf{I}}_{p(i)}
=
\trcorner{C,(k)}{\mathbf{I}}_{p(i)}
+
 (\tlcorner{i}{\mathbf{X}}_{p(i)}^{(k)})^{\top} 
 \; \trcorner{C,(k)}{\mathbf{I}}_{i} 
 \; \tlcorner{i}{\mathbf{X}}_{p(i)}^{(k)}$
\STATE \hspace{0.3cm} \textbf{end}
\STATE \hspace{0.3cm} $\tlcorner{0}{\mathbf{I}}_{\mathcal{G}}^{(k)} = \trcorner{C,(k)}{\mathbf{I}}_{0}$
\STATE \hspace{0.3cm} retrieve $\tlcorner{0}{\mathbf{r}}_{\mathcal{G}}^{(k)}$ from $\tlcorner{0}{\mathbf{I}}_{\mathcal{G}}^{(k)}$
\STATE \hspace{0.3cm} compute $\tlcorner{0}{\mathbf{X}}_{\mathcal{C}}^{(k)}$ via~\eqref{equ:center_root_adjoint_transformation}
\STATE \hspace{0.3cm} $\tlcorner{\mathcal{C}}{\mathbf{I}}_{\mathcal{G}}^{(k)}
= 
(\tlcorner{0}{\mathbf{X}}_{\mathcal{C}}^{(k)} )^{\top}
\tlcorner{0}{\mathbf{I}}_{\mathcal{G}}^{(k)} 
\tlcorner{0}{\mathbf{X}}_{\mathcal{C}}^{(k)}$
\STATE \hspace{0.0cm} \textbf{end}
\end{algorithmic}
\label{alg3}
\end{algorithm}


\subsection{Simplified Predictive Centroidal Dynamics for MPC}
\label{subsec:centroidal_dynamics}
\begin{figure}[!t]
	\vspace{6pt}
\includegraphics[trim={0cm 0cm 0cm 0cm},clip,width=\linewidth]{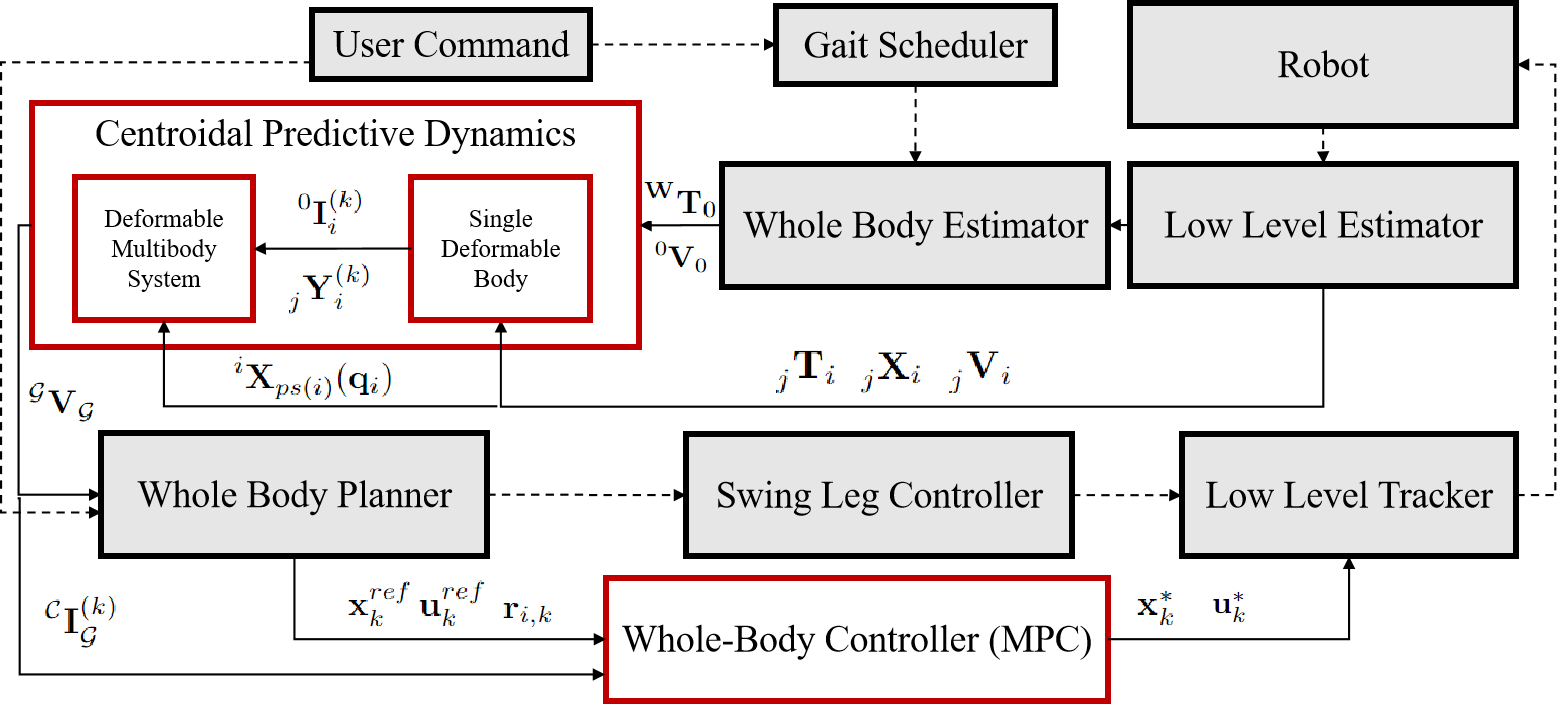}
	\vspace{-9pt}
	\caption{\textbf{Control Framework Diagram.} 
 This paper addresses two control components (shown in red blocks); the centroidal predictive dynamics and MPC-based whole-body controller. The other control components (grey-shaded) and the corresponding signal pattern (dashed-arrow) are beyond the scope of this work.  
 }\label{fig:control_framework}
	\vspace{0pt}
\end{figure}

We investigate the centroidal dynamics that can be exploited in MPC for locomotion (Fig.~\ref{fig:control_framework}). 
Many existing scenarios~\cite{di2018dynamic,grandia2019feedback} accept the potato model to approximate the robot trunk as a single rigid body and thus study the states of the body frame around the rigid body's centroid. 
In the case of the deformable multibody system, we can work on the states of the center-body frame $f_{\mathcal{C}}$. 

The continuous-time centroidal dynamics at each time step $t_k$ takes the state-space form~\cite{di2018dynamic} as
\begin{align}
& \dot{\mathbf{x}}_k = \mathbf{A}_{c,k} \mathbf{x}_k + \mathbf{B}_{c,k} \mathbf{u}_k\;,
\label{equ:continuous_time_centroidal_dynamics_state_space_form}
\\
& \mathbf{x}_k = [ \mathbf{\Theta}^\top, \mathbf{p}^\top, \boldsymbol{\omega}^\top, \dot{\mathbf{p}}^\top, g_z ]^\top
,\mathbf{u}_k = [ \mathbf{f}_1^\top, \cdots , \mathbf{f}_{N_f}^\top ]^\top\;,
\label{equ:continuous_time_centroidal_dynamics_control_def}
\end{align}
where the state vector $\mathbf{x}_k$ consists of the orientation $\mathbf{\Theta}$, the position $\mathbf{p}$, the angular velocity $\boldsymbol{\omega}$, and linear velocity $\dot{\mathbf{p}}$ around the centroid expressed in the world frame $f_{\mathbf{W}}$. 
State $\mathbf{x}_k$ also contains the vertical gravity component $g_z$ to include the gravitational effect. 
The orientation is defined in ZYX Euler angles as $\mathbf{\Theta} = [\phi, \theta, \psi]^\top$. 
The control vector $\mathbf{u}_k$ is the collection of all ground reactive forces $\mathbf{f}_i$. 

The assumption of small roll and pitch angles leads to the simplification of the state and control matrices in~\eqref{equ:continuous_time_centroidal_dynamics_state_space_form} as functions of yaw rotation $\mathbf{R}_z(\psi)$, centroid-foothold vector $\mathbf{r}_{i,k}$, and world-frame centroid inertia $\hat{\mathbf{I}}_k$~\cite{di2018dynamic}. 
The CCPDI tensor $\tlcorner{\mathcal{C}}{\mathbf{I}}_{\mathcal{G}}^{(k)}$ from Alg.~\ref{alg:deformable_multibody_centroidal_composite_inertia} approximates $\hat{\mathbf{I}}_k $ with its rotational inertia as
\begin{align}
\hat{\mathbf{I}}_k 
= 
\mathbf{R}_z(\psi) 
\tlcorner{\mathcal{C}}{\bar{\mathbf{I}}}_{\mathcal{G}}^{(k)} 
\mathbf{R}_z(\psi)^\top,
\label{equ:mpc_ccpdi_induce}
\end{align}
Further discretization of~\eqref{equ:continuous_time_centroidal_dynamics_state_space_form} results in the discrete-time centroidal dynamics at each time step $t_k$ as
\begin{align}
\mathbf{x}_{k+1} = \mathbf{A}_{d,k} \mathbf{x}_k + \mathbf{B}_{d,k} \mathbf{u}_k\;,
\label{equ:discrete_time_centroidal_dynamics_state_space_form}
\end{align}
which is used as the system dynamic constraint in the online convex linear MPC problem~\cite{di2018dynamic}.



\section{Results and Discussion}\label{sec:Results}

\begin{table}[!t]
\vspace{0pt}
    \caption{MPC Configuration Parameters} 
    \vspace{-9pt}
    \label{table:mpc_configuration_parameters}
    \begin{center}
    \renewcommand{\arraystretch}{1.5}
    \begin{tabular}{p{3.2cm}>{\centering}p{2.25cm}>{\centering}p{1.9cm}}
        \toprule
        \textbf{Parameters} & Symbol * 
        & Values  \\ 
        \midrule
        \midrule
        Postion Weights& $Q_{p,x}$, $Q_{p,y}$,$Q_{p,z}$& 1e-5,1e-5,2e3\\ 
        Velocity Weights& $Q_{v,x}$, $Q_{v,y}$,$Q_{v,z}$& 1e3,1e3,1e2\\  
        Euler Angle Weights& $Q_{e,x}$, $Q_{e,y}$,$Q_{e,z}$& 1e2,4e2,1e2\\
 Angular Velocity Weights& $Q_{e,x}$, $Q_{e,y}$,$Q_{e,z}$&1e1,1e1,1e1\\
 GRF Weights& $R_{f,x}$, $R_{f,y}$,$R_{f,z}$&1e-8,1e-8,1e-8\\
        \bottomrule
    \end{tabular}
    \end{center}
    \footnotesize{}
    \vspace{-18pt}
\end{table}

\subsection{Environment and Robot Setup}
The simulation models are established based on the physical prototype's design parameters 
and are deployed in Webots 2023b. 
We avoid position and velocity-level commands to narrow the gap between the simulated and the real systems. 
We only send computed force and torque commands to the target joints for less smooth but more realistic motions during the robot-environment interaction. 

The MPC weights (Table~\ref{table:mpc_configuration_parameters}) are selected for the simulation model with the rigid spine to stabilize the stepping with trot gait.
The weights and other MPC setups remain the same for the compliant model tests.  

For the spinal compliance, the spring constant and rest length are the major parameters of interest. 
When the spring is stiff and its rest length is longer than the spine length, it is always compressed and tensions the spine to its maximum length and is less likely to deform. 
Therefore, this paper investigates a much softer setup where the spring is less stiff~\cite{ye2021modeling} and its rest length is within the spine length bounds such that the spine is more likely to deform during locomotion. 

\begin{figure}[!t]
	\vspace{0pt}
	\centering
\includegraphics[trim={1.5cm 0cm 3.5cm 0cm},clip,width=\linewidth]{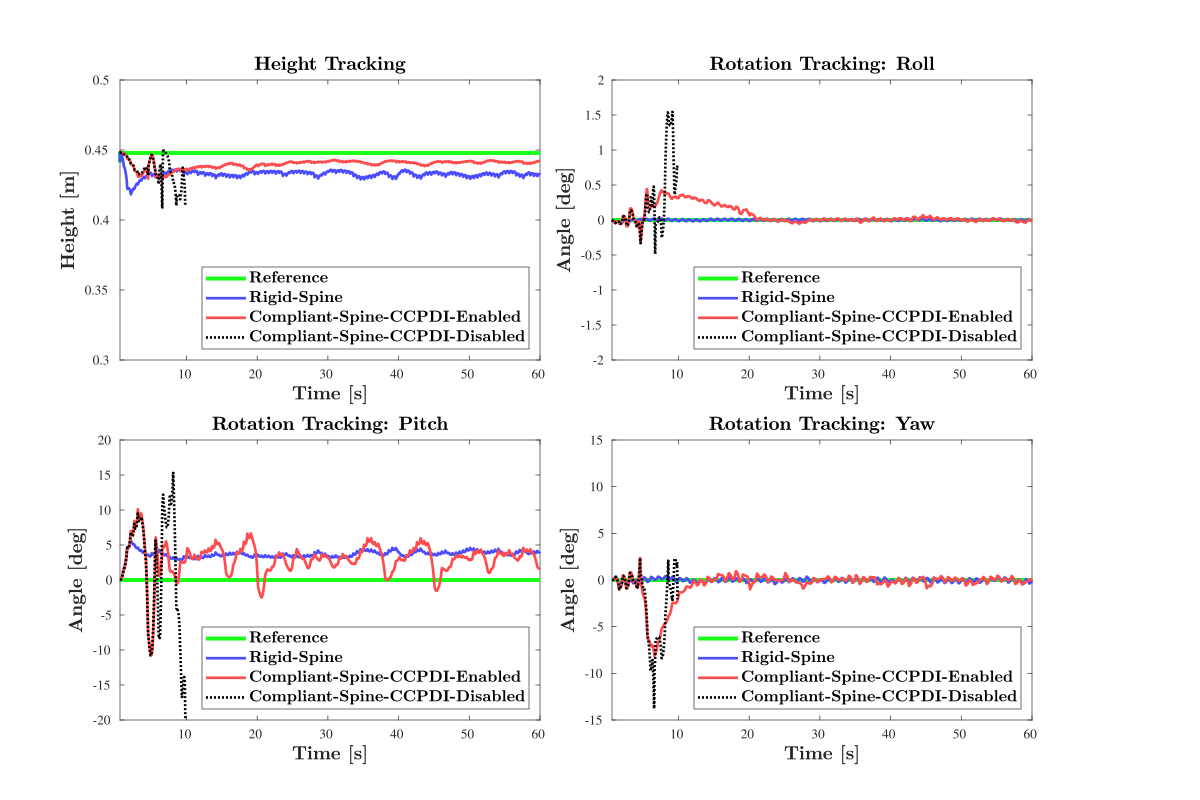}
	\vspace{-22pt}
	\caption{\textbf{Trajectory of Centroid States. } 
 Tracking performance of 1-minute trot stepping is investigated between the rigid (blue) and compliant models, and between CCPDI-enabled and CCPDI-disabled MPC settings for the compliant model in four directions: height (top-left), roll angle (top-right), pitch angle (bottom-left), and yaw angle (bottom-right). Spinal compliance is configured as $k_s=36 N/m$ and $l_{rest} = 0.180 m$. }
 \label{fig:results_states_trajectory_all}
	\vspace{-18pt}
\end{figure}

\begin{figure}[!t]
	\vspace{6pt}
\includegraphics[trim={3.3cm 0cm 2cm 0cm},clip,width=\linewidth]{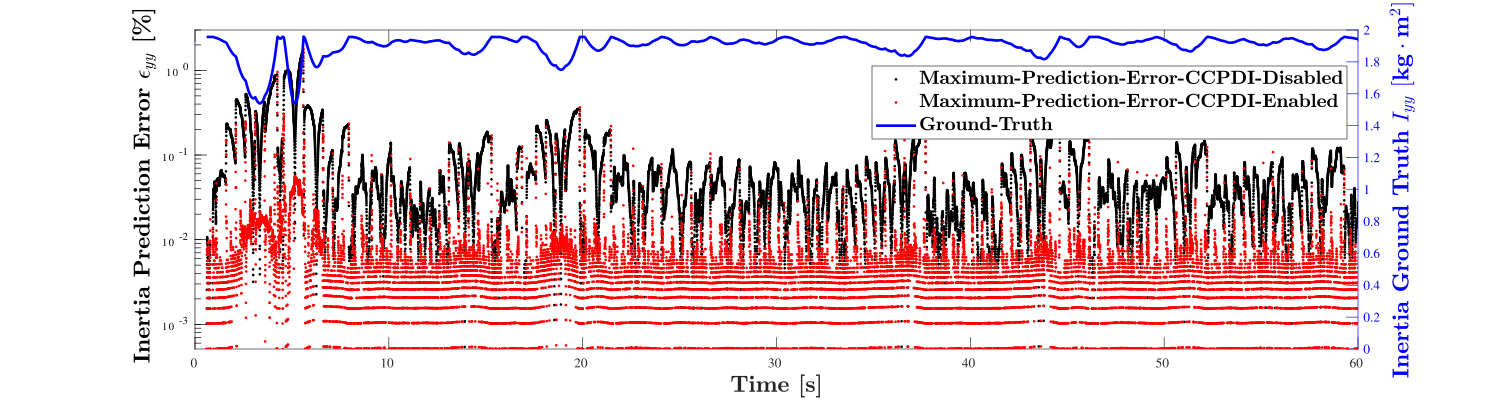}
	\vspace{-18pt}
	\caption{\textbf{Inertia Prediction with and without CCPDI. } 
 The maximum absolute error of predicted composite inertia in y-axis $\epsilon_{yy}$ is compared between CCPDI-enabled (red) and CCPDI-disabled (black) MPC settings for time varying composite inertia (blue).}
 \label{fig:results_prediction_error}
	\vspace{-15pt}
\end{figure}



\subsection{Trot Stepping: Rigid and Compliant Models}
The tests for dynamic gaits in this work concentrate on stepping with the trotting gait. 
Both rigid and compliant robots are tested with the same centroidal predictive dynamics module and MPC module (Fig.~\ref{fig:control_framework}). 

Figure~\ref{fig:results_states_trajectory_all} demonstrates the results of an instance comparable test between the rigid model with standard MPC and the compliant model with CCPDI-enabled MPC. 
Both models show similar tracking performance regarding height and roll angle. Yet, the compliant model experiences larger oscillation in pitch and yaw angles due to the morphological change of the spine.
Note that the CCPDI-enabled MPC on the compliant model eventually stabilizes the yaw angle and bounds the pitch angle oscillation.  


\subsection{Trot Stepping: CCPDI-enabled and CCPDI-disabled}
\begin{figure}[!t]
	\vspace{6pt}
\includegraphics[trim={3cm 0cm 2cm 0cm},clip,width=\linewidth]{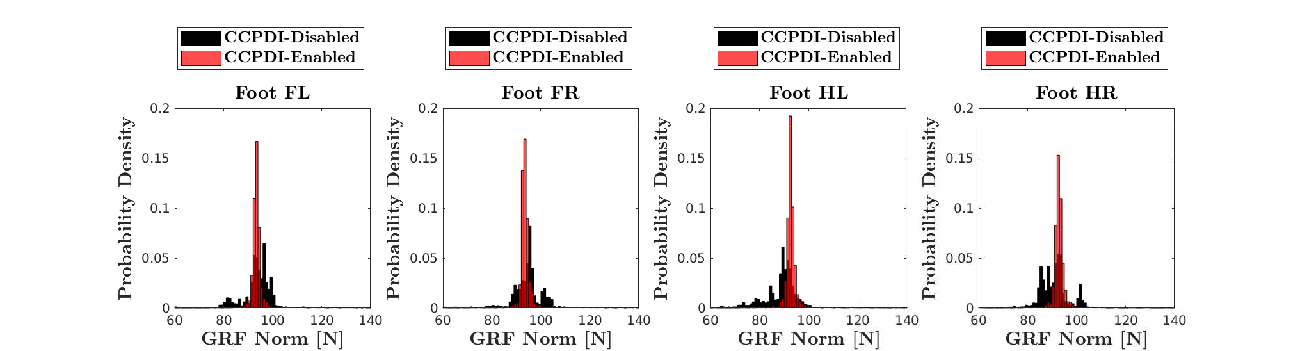}
	\vspace{-15pt}
	\caption{\textbf{Distribution of Planned GRF Norm with and without CCPDI. } 
 GRF's $l_2$ norm distribution is analyzed between CCPDI-enabled (red) and CCPDI-disabled (black) MPC settings across legs.}
 \label{fig:results_grf_distribution_ccpdi_enabled_vs_disabled}
	\vspace{-10pt}
\end{figure}

Equation~\eqref{equ:mpc_ccpdi_induce} explains how CCPDI tensors at $k^{th}$ step are embedded into the MPC's system dynamic constraints. 
CCPDI-disabled means treating the compliant model as a rigid model throughout the prediction horizon. 
In other words, only the first CCPDI tensor is used for MPC.  
Figure~\ref{fig:results_states_trajectory_all} illustrates the trajectory of the composite centroid states for the same compliant model with and without the CCPDI tensors. 
All other MPC settings remain the same between these two cases. 
The CCPDI-disabled MPC fails to stabilize the robot after 9 seconds, with a major divergence observed in the pitch angle.
The absolute prediction error of the principal composite inertia along the y-axis $\epsilon_{yy}^{(k)}(t):=|(\tlcorner{0}{\mathbf{I}}_{\mathcal{G},yy}^{(k)}(t)-\tlcorner{0}{\mathbf{I}}_{\mathcal{G},yy}^{(0)}(t+k\Delta t))/\tlcorner{0}{\mathbf{I}}_{\mathcal{G},yy}^{(0)}(t+k\Delta t)|$ illustrates the prediction accuracy when the composite inertia changes dynamically and $\epsilon_{yy}(t):=\text{max}\{\epsilon_{yy}^{(k)}(t)\}$ denotes the worst case within the same prediction horizon. 
Figure~\ref{fig:results_prediction_error} shows that CCPDI offers a better average inertia prediction accuracy than without CCPDI by around 10 times. 
Figure~\ref{fig:results_grf_distribution_ccpdi_enabled_vs_disabled} demonstrates the distribution of planned GRFs from MPC with different CCPDI options. 
It is observed that the CCPDI-enabled MPC guides the GRF solutions closer to and symmetric around the heuristic of evenly distributed for whole-body weight compensation and balance. 
On the other hand, CCPDI-disabled MPC produces more diverging solutions that could undermine overall gait stabilization.  


\begin{figure}[!t]
	\vspace{0pt}
	\centering
\includegraphics[trim={0cm 0cm 0cm 0cm},clip,width=0.5\linewidth]{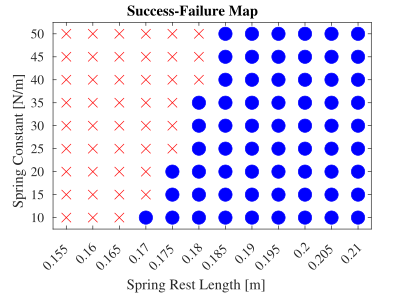}
	\vspace{-2pt}
	\caption{\textbf{Performance Map of CCPDI-Enabled MPC. } MPC performance is evaluated across compliant configurations as $k_s = 10 \sim 50 N/m$ and $l_{rest} = 0.155 \sim 0.21 m$. Successful cases are shown with blue dots, and failure cases are marked by red crosses.}
 \label{fig:performance_map}
	\vspace{-18pt}
\end{figure}

\subsection{Effect of Compliance }
We investigate various compliant configurations to test the robustness of the proposed method. 
Figure~\ref{fig:performance_map} shows the success-and-failure map across the chosen parameters. 
The CCPDI-enabled MPC is robust to different spring stiffness when the spring rest length is longer than 0.18 $m$, yet substantially underperforms once the spring rest length is shorter than 0.175 $m$. 
This observation offers engineering insights that the proposed method may be suitable for more heavily spring-loaded systems. 

\section{Conclusions}

The paper contributes to dynamic gait stabilization for a quadrupedal robot with embodied compliance. 
The proposed approach develops a unified modeling technique to approximate both rigid and compliant bodies for their deformation. 
A deformable multibody system is established with the computation of the centroidal composite predictive deformed inertia (CCPDI) tensor and the integration with MPC regulation.
The simulation results suggest the feasibility of the proposed framework in stabilizing the trot stepping for both rigid and compliant models. 
Future directions of research include integration of CCPDI in overall control framework and verification in our physical prototype robot. 

\newpage




\end{document}